\documentclass[10pt,twocolumn,letterpaper]{article}

\usepackage[pagenumbers]{cvpr} 

\definecolor{cvprblue}{rgb}{0.21,0.49,0.74}
\usepackage[pagebackref,breaklinks,colorlinks,allcolors=cvprblue]{hyperref}

\def\paperID{5576} 
\def\confName{CVPR}
\def\confYear{2025}

\title{SALAD: Skeleton-aware Latent Diffusion \\ for Text-driven Motion Generation and Editing}

\author{Seokhyeon Hong \quad Chaelin Kim \quad Serin Yoon \quad Junghyun Nam \quad Sihun Cha \quad Junyong Noh \\
Visual Media Lab, KAIST\\
{\tt\small \{ghd3079, chaelin.kim, serinyoon, ys4990, chacorp, junyongnoh\}@kaist.ac.kr}
}

\usepackage{amsmath, mathtools, amssymb}
\usepackage{multirow, array}
\usepackage{arydshln}
\usepackage{ulem}
\usepackage{graphics}
\usepackage{makecell}

\usepackage{inconsolata}
\usepackage{algorithm, algorithmic}

\newcommand{\ub}[1]{\underline{\textbf{#1}}}

\newcommand{\vfunc}[2]{\mathrm{#1}(\mathbf{#2})} 
\newcommand{\loss}[1]{\mathcal{L}_\mathrm{#1}}
\newcommand{\wloss}[1]{\lambda_\mathrm{#1}\mathcal{L}_\mathrm{#1}}
\newcommand{\velocity}{\mathbf{v}}
\newcommand{\sample}{\mathbf{x}}
\newcommand{\noise}{\boldsymbol{\epsilon}}
\newcommand{\zlt}{\mathbf{z}^\mathit{l}_\mathit{t}}
\newcommand{\zt}{\mathbf{z}_\mathit{t}}
\newcommand{\denoiser}[1]{\mathbf{v}_\theta(#1)}

\newcommand{\bestmetric}[2]{\color{red}{${#1}^{\pm#2}$}}
\newcommand{\secondmetric}[2]{\color{blue}{${#1}^{\pm#2}$}}
\newcommand{\metric}[2]{${#1}^{\pm#2}$}
\newcommand{\ca}{\centering\arraybackslash}

\usepackage[capitalize]{cleveref}
\crefname{section}{Sec.}{Secs.}
\Crefname{section}{Section}{Sections}
\Crefname{table}{Table}{Tables}
\crefname{table}{Tab.}{Tabs.}

\begin{document}
\maketitle
\begin{abstract}
Text-driven motion generation has advanced significantly with the rise of denoising diffusion models.
However, previous methods often oversimplify representations for the skeletal joints, temporal frames, and textual words, limiting their ability to fully capture the information within each modality and their interactions.
Moreover, when using pre-trained models for downstream tasks, such as editing, they typically require additional efforts, including manual interventions, optimization, or fine-tuning.
In this paper, we introduce a skeleton-aware latent diffusion~(SALAD), a model that explicitly captures the intricate inter-relationships between joints, frames, and words.
Furthermore, by leveraging cross-attention maps produced during the generation process, we enable attention-based zero-shot text-driven motion editing using a pre-trained SALAD model, requiring no additional user input beyond text prompts.
Our approach significantly outperforms previous methods in terms of text-motion alignment without compromising generation quality, and demonstrates practical versatility by providing diverse editing capabilities beyond generation.
Code is available at {\href{https://seokhyeonhong.github.io/projects/salad/}{project page}}.
\end{abstract}
\vspace{-1em}
\section{Introduction}
\label{sec:intro}
Character animation is a crucial component in various computer graphics and vision applications, including games, films, and interactive media.
Traditional methods, such as keyframing and motion capture, typically require extensive manual effort, which are time-consuming and expensive.
Recently, deep generative models have been introduced to mitigate these challenges.
In particular, diffusion models have shown promising results in text-to-motion generation, enabling intuitive and efficient animation workflows.

Despite these advancements, fully capturing the intricate relationships among frames, body parts, and textual descriptions remains a complex challenge in text-driven motion generation.
Previous approaches typically represent a pose as a single vector and primarily focus on temporal relationships between poses across frames, while neglecting spatial interactions between skeletal joints.
Furthermore, these models often compress a sentence into a single vector for conditioning, which can overlook the important nuances of word-level variations.
This over-simplification of interactions among skeletal, temporal, and textual components may lead to missing details in the generated results, underscoring the need for models that faithfully capture these complex dependencies.

On the other hand, learning meaningful representations is a key factor to enable zero-shot downstream tasks with pre-trained models.
For example, zero-shot text-driven image and video editing methods~\cite{hertz2022prompt2prompt, alaluf2024cross, cao2023masactrl, tumanyan2023plugnplay, qi2023fatezero, liu2024videop2p, guo2024focus} leverage attention modulation from pre-trained models for intuitive manipulation.
Unfortunately, motion generation models typically lack such interpretable and manipulatable intermediate representations.
This limitation stems from the over-simplification of motion and text features mentioned above, which restricts rich interactions between them.
Although some methods exploit pre-trained models with manual masks, optimization, or fine-tuning~\cite{tevet2023mdm, kim2023flame, karunratanakul2024dno, karunratanakul2023gmd, wan2023tlcontrol, xie2023omnicontrol}, these methods require additional effort, time, and cost to achieve desired results.
Therefore, developing interpretable representations during motion synthesis could improve versatility and flexibility for downstream tasks beyond generation, such as zero-shot motion editing.

In this paper, we propose a \ub{s}keleton-\ub{a}ware \ub{la}tent \ub{d}iffusion, which we call SALAD, for text-driven motion generation within a skeleto-temporally structured latent space.
We first train a variational autoencoder~(VAE)~\cite{kingma2013vae} to construct a motion latent space that decouples spatial and temporal dimensions.
To this end, we employ skeleto-temporal convolution layers to facilitate information exchange between adjacent joints and frames.
We also employ skeleto-temporal pooling layers to produce a compact representation by reducing the dimensionality, which can reduce the computational complexity during the sampling process of the diffusion model.
Subsequently, we train a diffusion model within this latent space to generate motion features conditioned on text.
To train the denoiser, we leverage both skeletal and temporal attention blocks, which can effectively enable skeleto-temporal coherence.
Additionally, we employ cross-attention to capture fine-grained interactions between individual words and the skeleto-temporal units of the motion latents.

Beyond motion generation, we introduce an attention-based zero-shot, text-driven motion editing method that leverages the pre-trained SALAD model, requiring no additional optimization or fine-tuning.
We demonstrate that the intermediate cross-attention maps of SALAD capture the relationship between text and motion features.
By modulating these cross-attention maps, we enable text-driven motion editing that allows users to edit generated motion through text input alone in a training-free manner.
We also propose a novel attention modulation method for motion-specific tasks, demonstrating the practical potential of our approach.

In summary, our contributions are as follows:
\begin{itemize}
    \item We propose SALAD, a novel skeleton-aware latent diffusion model for text-driven motion generation within a skeleto-temporally structured latent space.
    \item We interpret the intermediate representations in the generation process, allowing for a clear understanding of the relationship between text inputs and generated motions. 
    \item We present an attention-based zero-shot text-driven motion editing method using cross-attention modulation in generative models.
\end{itemize}
\section{Related Work}
\label{sec:related}

\subsection{Text-driven Motion Generation}
With the rise of deep neural networks, generative models for human motion synthesis based on natural language descriptions have been extensively explored.
Early studies focused on generating motions from action labels, such as \textit{throw} and \textit{kick}~\cite{petrovich2021action, guo2020action2motion}.
More advanced methods projected both text and motion features into a shared latent space, enabling the generation of fixed motion sequences from identical input text~\cite{ahuja2019language2pose}.
To increase the diversity of generated results, T2M~\cite{guo2022t2m} employed a temporal VAE that can generate varied motion sequences that align with the given text input.
Subsequent studies leveraged large pre-trained text encoders, such as CLIP~\cite{radford2021clip}, to improve textual understanding for motion generation~\cite{petrovich2022temos, athanasiou2022teach, tevet2022motionclip}.
Auto-regressive models that behave like language models have also been proposed, which first encode motion features into discrete tokens through vector quantization~\cite{van2017vqvae} and then auto-regressively predict the next motion token given the text prompt~\cite{guo2022tm2t, jiang2023motiongpt, zhang2023t2mgpt, zhang2024motiongpt, zhong2023attt2m, guo2024momask, zou2024parco}.
Recently, denoising diffusion models~\cite{ho2020denoising, sohl2015deep} have become the dominant framework for generative models, prompting text-driven motion generation methods to adopt diffusion models as their backbone~\cite{tevet2023mdm, chen2023mld, zhang2023remodiffuse, zhang2024motiondiffuse, wang2023fg, lou2023diversemotion, kim2023flame}.
In this study, we build our framework on the diffusion model formulation and enhance the performance of text-driven motion generation through explicit modeling of the relationships among joints, frames, and words.
Moreover, we introduce an attention-based zero-shot text-driven editing method, leveraging our pre-trained SALAD model as a  generator, demonstrating a novel exploration beyond mere generation.

\subsection{Skeleton-aware Processing of Motion Data}
Because motion data consists of two separate skeletal and temporal dimensions, some methods have been developed to respect each dimension when processing the motion data.
Yan et al.~\cite{yan2018spatial} presented spatial-temporal graph convolution networks that can learn both the spatial and temporal patterns of dynamic skeletons.
Skeleton-aware networks~\cite{aberman2020skeleton} employed skeleto-temporal convolution networks and topology-preserving skeletal pooling layers to construct a shared motion latent space, enabling deep motion retargeting across homeomorphic skeletons.
Motion style transfer has also been enabled in a part-wise manner by employing skeleto-temporal convolution modules~\cite{park2021diverse, jang2022motionpuzzle}.
In the realm of text-to-motion generation, ParCo~\cite{zou2024parco} and AttT2M~\cite{zhong2023attt2m} utilized the skeleto-temporal network architecture during the motion quantization process.
In this work, we adopt a similar approach by encoding motion features using skeleto-temporal convolution networks and pooling layers to derive an effective skeleton-aware motion latent space with compact dimensionality.
Additionally, we facilitate the interaction between this skeleton-aware motion latent space and textual features during the text-conditional generation process.

\begin{figure*}[ht]
  \centering
   \includegraphics[width=\linewidth]{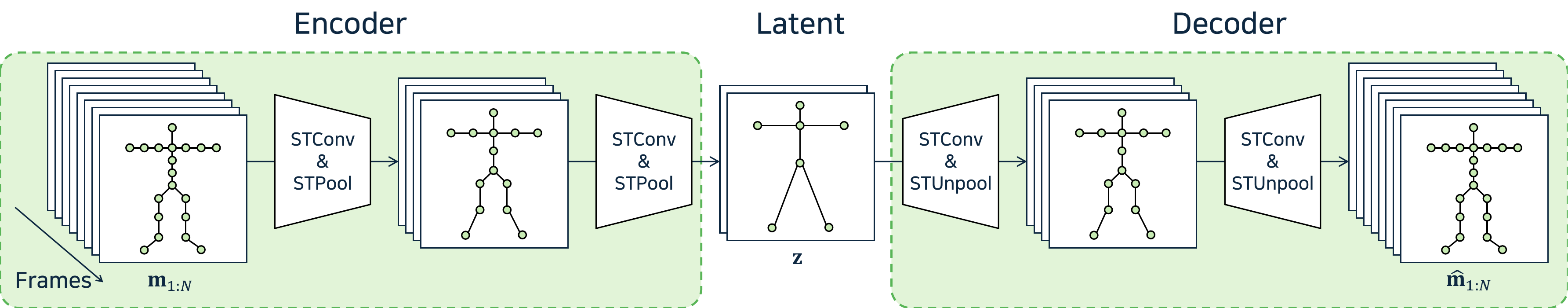}
   \vspace{-1.5em}
   \caption{Architecture of the skeleto-temporal VAE network. The encoder maps motion features into a skeleto-temporal latent space, and the decoder restores the skeleto-temporal latent variables into motion features.}
   \label{fig:vae}
   \vspace{-1.5em}
\end{figure*}
\subsection{Zero-shot Editing with Diffusion Models}
By leveraging the generative capabilities of diffusion models, zero-shot editing methods have been proposed.
SDEdit~\cite{meng2021sdedit} first adds noise to the input data for a few diffusion timesteps, then denoises the noisy data with an editing condition to achieve high-fidelity zero-shot editing by leveraging a pre-trained diffusion model.
Prompt-to-Prompt~\cite{hertz2022prompt2prompt} verified that cross-attention maps establish connections between word tokens and the spatial layout of the generated image.
In addition, it introduced attention modulation methods that can be utilized during the generation process, enabling the generation of high-fidelity editing results in a zero-shot manner.
Null-text inversion~\cite{mokady2023null} extended this method to real-images by improving the diffusion inversion method.
Similarly, image and video editing methods via attention modulation have been extensively studied~\cite{alaluf2024cross, cao2023masactrl, tumanyan2023plugnplay, qi2023fatezero, liu2024videop2p, guo2024focus}.
Motion diffusion models often employ joint masks~\cite{tevet2023mdm, kim2023flame} or optimization~\cite{xie2023omnicontrol, karunratanakul2024dno, karunratanakul2023gmd} to enable motion editing using pre-trained diffusion models, although they require user intervention or additional computation for the desired goal.
On the other hand, CoMo~\cite{huang2024controllable} achieved zero-shot text-driven motion editing using large language models to manipulate semantic pose tokens.
We demonstrate that the intermediate cross-attention maps of our SALAD model can capture the relationship between motion and text, similarly to image diffusion models. Furthermore, by modulating these cross-attention maps, we introduce an attention-based zero-shot text-driven motion editing method that eliminates the need for manual masks, fine-tuning, or optimization.

\section{Method}
\label{sec:method}
Our goal is to generate a motion sequence $\mathbf{m}_{1:N}$ conditioned on a text prompt $c$, where $N$ denotes the number of frames. 
We first construct a skeleton-aware motion latent representation that captures both skeletal and temporal dynamics of a motion sequence~(\cref{sec:method-vae}).
Using this motion representation, we train a diffusion model by modeling the complex interactions between skeletal joints, temporal frames, and textual descriptions~(\cref{sec:method-denoiser}).
Finally, we extend our model for zero-shot text-driven motion editing by modulating cross-attention maps between text and motion~(\cref{sec:method-editing}).
Additional details on the network architectures will be featured in the supplementary material.

\begin{figure*}[t]
  \centering
   \includegraphics[width=\linewidth]{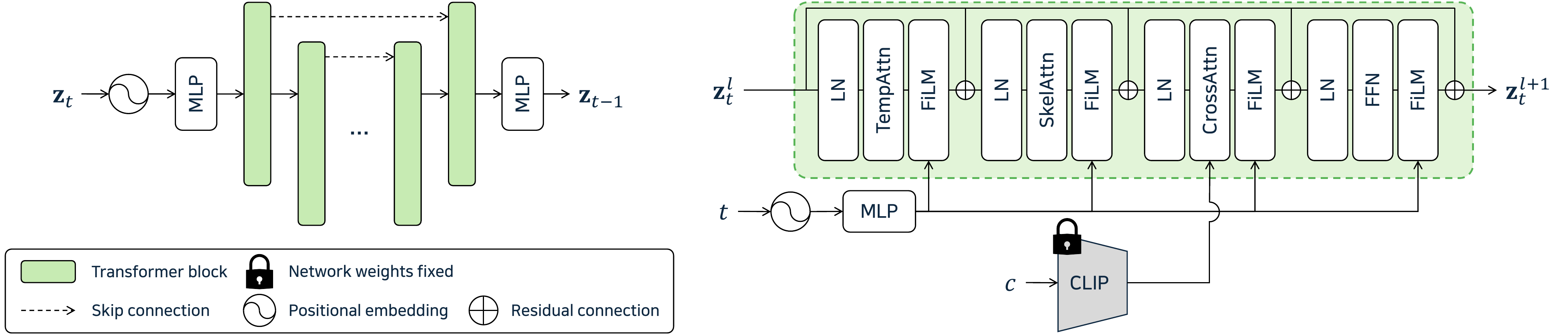}
   \vspace{-1.5em}
   \caption{(Left) Overall network architecture of the denoiser. (Right) The architecture of each transformer block consisting of TempAttn, SkelAttn, CrossAttn, and FFN, along with the FiLM following each module.}
   \label{fig:denoiser}
   \vspace{-1.5em}
\end{figure*}
\subsection{Skeleton-aware VAE}
\label{sec:method-vae}
To derive skeleton-aware motion features, we train a VAE composed of skeleto-temporal convolution networks and pooling layers~\cite{yan2018spatial, aberman2020skeleton}, as illustrated in~\cref{fig:vae}.
The core module is the skeleto-temporal convolution~(STConv), which decouples joint and frame dimensions while facilitating information exchange between adjacent components within each dimension.
While the skeleto-temporal separation of motion sequences allows for sophisticated modeling of motion data, it introduces an additional dimension for skeletal joints, unlike previous text-to-motion generation methods that treat a pose as a single vector.
This expanded data space increases the risk of curse of dimensionality and leads to longer computation times for diffusion models, which rely on solving stochastic differential equations within this higher-dimensional data space.
To address this issue, we apply skeleto-temporal pooling~(STPool) layers in the encoder to reduce the latent space dimensionality, effectively compressing the representation.
In the decoder, skeleto-temporal unpooling~(STUnpool) layers are used to reconstruct the compressed latent features back to raw motion.

\noindent{\textbf{Encoder.}}
We separate each pose vector of $\mathbf{m}_{1:N}$ into joint-wise features and propagate them through joint-wise multi-layer perceptron~(MLP) layers, resulting in $\mathbf{h} \in \mathbb{R}^{N \times J \times D}$, where $J$ denotes the number of joints and $D$ denotes the latent dimension.
Subsequently, we apply STConv to $\mathbf{h}$ to facilitate information exchange between adjacent frames and joints:
\begin{equation}
    \vfunc{STConv}{h} \coloneq \vfunc{SkelConv}{h} + \vfunc{TempConv}{h},
\end{equation}
where SkelConv is a graph convolution network over the joint dimension, while TempConv is a 1D convolution network over the frame dimension.

We apply the STPool operator to the output of the STConv layers to reduce its dimensionality by pooling over joints and frames:
\begin{equation}
    \vfunc{STPool}{h} \coloneq \vfunc{TempPool}{\vfunc{SkelPool}{h}},
\end{equation}
where SkelPool represents the pooling layer over the joint dimension while TempPool represents the pooling layer over the temporal dimension.
Specifically, SkelPool aggregates adjacent joints while preserving its skeletal topology, resulting in fewer joints with a homeomorphic structure.
In contrast, TempPool performs 1D pooling over frames.
These pooling operations are commutative because they operate within independent dimensions.

Consequently, we obtain the skeleto-temporally compressed latent vector $\mathbf{z} \in \mathbb{R}^{N' \times J' \times D}$, where $N' < N$ and $J' < J$.
This representation effectively encodes the skeleton-aware dynamics of the motion sequence within a lower dimension.
Notably, we maintain 7 atomic joints for this motion latent: root, spine, head, both arms, and both legs ($J'=7$).

\noindent{\textbf{Decoder.}}
To reconstruct $\mathbf{z}$ into raw motion features, we employ STConv and STUnpool layers in the decoder.
The decoder mirrors the encoder, facilitating information exchange between adjacent joints and frames using STConv layers and increasing the skeleto-temporal resolution using STUnpool layers.
Finally, the joint-wise features are processed by their respective joint-wise MLP layers, resulting in the reconstructed motion feature $\hat{\mathbf{m}}_{1:N}$.

\noindent{\textbf{Training.}}
The VAE is trained with the following objectives:
\begin{equation}
    \loss{VAE} = \loss{\mathbf{m}} + \wloss{pos} + \wloss{vel} + \wloss{kl},
\end{equation}
where $\loss{\mathbf{m}}, \loss{pos}, \loss{vel}$ are L1 reconstruction loss terms of the motion features, joint positions, and joint velocities, while $\loss{kl}$ is the Kullback–Leibler~(KL) divergence regularization that encourages a structured latent space.

\subsection{Skeleton-aware Denoiser}
\label{sec:method-denoiser}
\noindent\textbf{Network Architecture.}
Given the skeleton-aware motion latent vector $\mathbf{z} \in \mathbb{R}^{N' \times J' \times D}$, we train a diffusion model that denoises the noised latent vector conditioned on text prompt $c$.
As shown in~\cref{fig:denoiser}, our denoiser consists of a positional embedding, MLP-based encoder and decoder, and $L$ transformer-based layers.
Each transformer layer is composed of a temporal attention~(TempAttn), skeletal attention~(SkelAttn), cross-attention~(CrossAttn), and feed-forward network~(FFN).
Each component involves a residual connection~\cite{he2016resnet}, layer normalization~(LN)~\cite{ba2016layer}, and a feature-wise linear modulation~(FiLM)~\cite{perez2018film} that modulates intermediate features based on a diffusion timestep.
We also incorporate skip connection~\cite{ronneberger2015unet} within the stack of transformer layers.

The input motion latent at diffusion timestep $t$ given to the $l$-th transformer layer, denoted as $\zlt$, is processed by three consecutive attention blocks: temporal, skeletal, and cross-attention layers.
The purpose of this architecture design is to model frame-wise interactions through TempAttn, joint-wise interactions through SkelAttn, and motion-text interactions through CrossAttn.
As a result, $\zlt$ is consecutively updated at each layer as follows:
\begin{gather}
    \zlt \leftarrow \zlt + \vfunc{FiLM}{\vfunc{TempAttn}{\vfunc{LN}{\zlt}}}, \\
    \zlt \leftarrow \zlt + \vfunc{FiLM}{\vfunc{SkelAttn}{\vfunc{LN}{\zlt}}}, \\
    \zlt \leftarrow \zlt + \vfunc{FiLM}{\vfunc{CrossAttn}{\vfunc{LN}{\zlt}, \vfunc{CLIP}{\mathit{c}}}}.
\end{gather}
For the CLIP text encoder~\cite{radford2021clip} used in cross-attention, we use a pre-trained model and freeze its weights during training.

\noindent{\textbf{Diffusion Parametrization.}}
\let\oldemptyset\emptyset
\let\emptyset\varnothing
For diffusion parametrization, our model predicts the diffusion velocity~\cite{salimans2022vpred}, denoted as $\velocity$, as follows:
\begin{equation}
    \velocity_{t} = \alpha_{t}\noise - \sigma_{t}\sample,
\end{equation}
where $\noise$ and $\sample$ represent the noise and sample, respectively, while $\alpha_t$ and $\sigma_t$ are noise scheduling parameters at diffusion timestep $t$.
While $\noise$- and $\sample$-prediction are widely used in diffusion models, $\velocity$-prediction combines both parametrizations, implicitly balancing the information contributed by each component.
This improves the stability of generation, especially at high noise levels in which the relationship between  $\noise$ and $\sample$ becomes disrupted.

\noindent{\textbf{Training and Inference.}}
The denoiser is trained to predict the diffusion velocity $\velocity_{t}$ at each diffusion timestep $t$.
The training objective for the denoiser is defined as follows:
\begin{equation}
    \loss{denoiser} = \lVert \hat{\mathbf{v}}_t - \mathbf{v}_t \rVert^2_2,
\end{equation}
where $\hat{\mathbf{v}}_t$ represents the predicted diffusion velocity by the denoiser.
During training, we randomly drop the text condition with a probability of $p_\mathrm{uncond}$, enabling unconditional generation when $c$ is a null text, which is represented as $\emptyset$.

To the trained denoiser, we apply classifier-free guidance~(CFG)~\cite{ho2022cfg} for text-conditioned generation:
\begin{equation}
    \hat{\mathbf{v}}_\theta (\zt, t, c) \coloneq \denoiser{\zt, t, \emptyset} + w\left( \denoiser{\zt, t, c} - \denoiser{\zt, t, \emptyset} \right).
\end{equation}
Here, $\mathbf{v}_\theta$ denotes the pre-trained denoiser, $w$ is the CFG weight, and $\hat{\mathbf{v}}_\theta$ is the modified velocity used during the denoising process.
Additionally, we employ the DDIM sampling formulation~\cite{song2020ddim} during inference, which effectively reduces the number of sampling steps without compromising generation quality.

\subsection{Zero-shot Text-driven Motion Editing}
\label{sec:method-editing}
\begin{figure}[t]
  \centering
   \includegraphics[width=\linewidth]{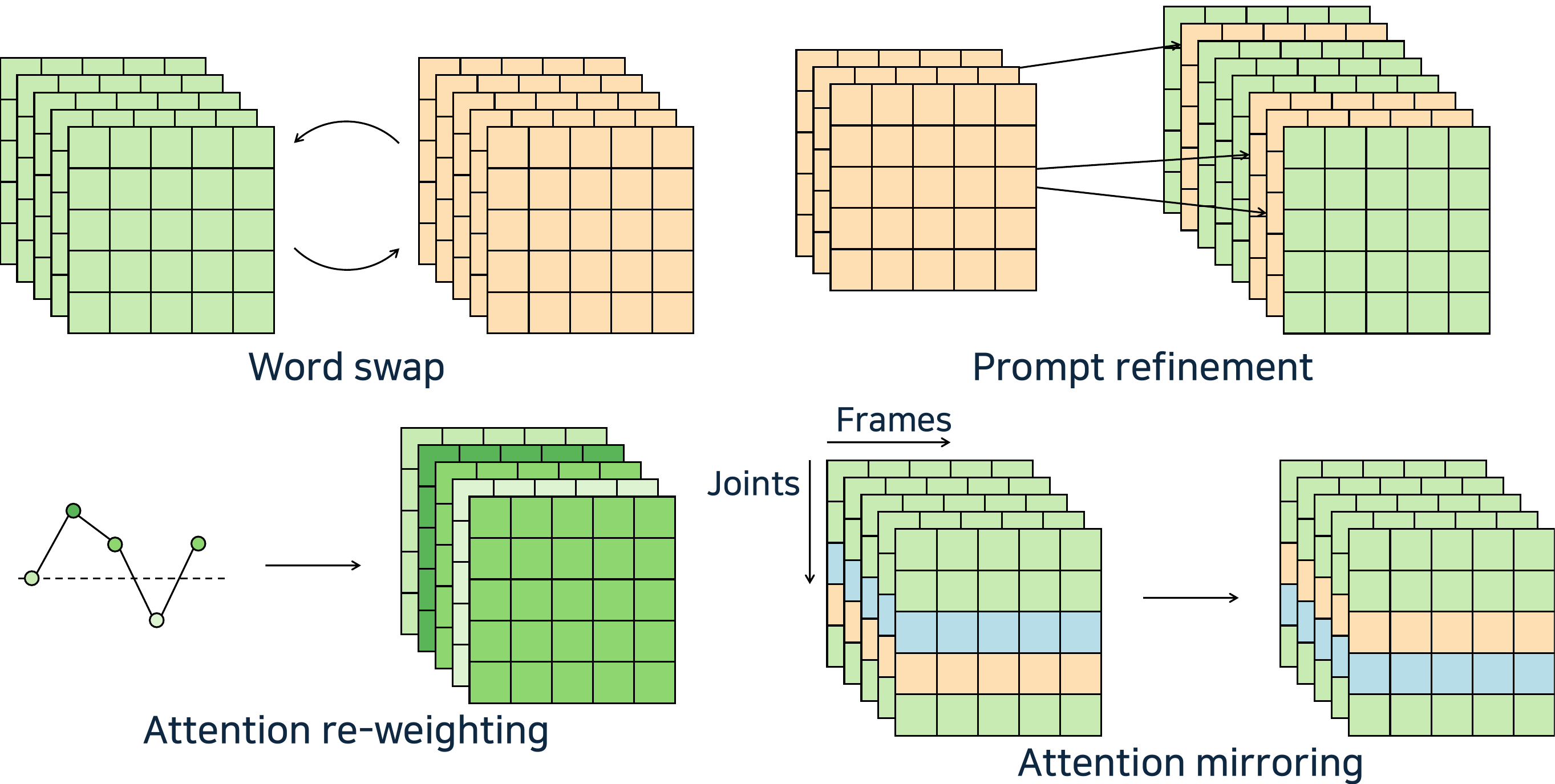}
   \vspace{-2em}
   \caption{Attention modulation methods applied to the cross-attention maps to enable text-driven motion editing.}
   \label{fig:attn_modulation}
   \vspace{-1.5em}
\end{figure}
In this section, we describe the attention modulation methods that enable zero-shot text-driven motion editing with a pre-trained SALAD model.
We introduce four distinct modulation strategies applied to the text-motion cross attention maps as shown in~\cref{fig:attn_modulation}: word swap, prompt refinement, attention re-weighting, and attention mirroring.
The first three are adapted from the Prompt-to-Prompt framework~\cite{hertz2022prompt2prompt}, while the attention mirroring serves as a validation method to assess whether the cross-attention map contributes to the activation of spatio-temporal dynamics in the motion generation process.
Word swap exchanges attention maps between a source and target prompt, enabling the model to leverage the attention context of the source text when generating output for the target text.
Prompt refinement enhances the original attention map by integrating additional attention maps for appended word tokens, providing richer semantic information compared to the base prompt.
Attention re-weighting adjusts the attention values assigned to specific user-selected words, either amplifying or diminishing their influence on the generated motion.
Lastly, attention mirroring swaps attention values between symmetrical body parts, such as the left and right arms, to produce mirrored motions.
These methods collectively empower the generative capacity of our model to produce dynamic and contextually relevant motion edits based solely on text prompts. 
For more details on these modulation functions, please refer to the supplementary material.

\begin{table*}[ht]
    \centering
    \resizebox{0.8\linewidth}{!}
    {
        \begin{tabular}{c>{\ca}c>{\ca}c>{\ca}c>{\ca}c>{\ca}c>{\ca}c>{\ca}c}
            \toprule
            \multirow{2}{*}{Methods} & \multicolumn{3}{c}{R-Precision ↑} & \multirow{2}{*}{FID ↓} & \multirow{2}{*}{MM-Dist ↓} & \multirow{2}{*}{Diversity →} & \multirow{2}{*}{MultiModality ↑} \\
            & Top-1 & Top-2 & Top-3 &  &  &  &  \\
            
            \midrule
            
            Real motion & \metric{0.511}{.003} & \metric{0.703}{.003} & \metric{0.797}{.002} & \metric{0.002}{.000} & \metric{2.974}{.008} & \metric{9.503}{.065} & - \\
            
            \midrule
            
            T2M~\cite{guo2022t2m} & \metric{0.457}{.002} & \metric{0.639}{.003} & \metric{0.740}{.003} & \metric{1.067}{.002} & \metric{3.340}{.008} & \metric{9.188}{.002} & \metric{2.090}{.083} \\
            AttT2M~\cite{zhong2023attt2m} & \metric{0.499}{.003} & \metric{0.690}{.002} & \metric{0.786}{.002} & \metric{0.112}{.006} & \metric{3.038}{.007} & \metric{9.700}{.090} & \secondmetric{2.452}{.051} \\
            ParCo~\cite{zou2024parco} & \metric{0.515}{.003} & \metric{0.706}{.003} & \metric{0.801}{.002} & \metric{0.109}{.005} & \secondmetric{2.927}{.008} & \secondmetric{9.576}{.088} & \metric{1.382}{.060} \\
            MoMask~\cite{guo2024momask} & \secondmetric{0.521}{.002} & \secondmetric{0.713}{.002} & \secondmetric{0.807}{.002} & \bestmetric{0.045}{.002} & \metric{2.958}{.008} & - & \metric{1.241}{.040} \\ \hdashline
            MDM~\cite{tevet2023mdm} & \metric{0.320}{.005} & \metric{0.498}{.004} & \metric{0.611}{.007} & \metric{0.544}{.044} & \metric{5.566}{.027} & \bestmetric{9.559}{.086} & \bestmetric{2.799}{.072} \\
            MLD~\cite{chen2023mld} & \metric{0.481}{.003} & \metric{0.673}{.003} & \metric{0.772}{.002} & \metric{0.473}{.013} & \metric{3.196}{.010} & \metric{9.724}{.082} & \metric{2.413}{.079} \\
            MotionDiffuse~\cite{zhang2024motiondiffuse} & \metric{0.491}{.001} & \metric{0.681}{.001} & \metric{0.782}{.001} & \metric{0.630}{.001} & \metric{3.113}{.001} & \metric{9.410}{.049} & \metric{1.553}{.042} \\
            ReMoDiffuse~\cite{zhang2023remodiffuse} & \metric{0.510}{.005} & \metric{0.698}{.006} & \metric{0.795}{.004} & \metric{0.103}{.004} & \metric{2.974}{.016} & \metric{9.018}{.075} & \metric{1.795}{.043} \\
            
            \midrule
            
            SALAD~(Ours) & \bestmetric{0.581}{.003} & \bestmetric{0.769}{.003} & \bestmetric{0.857}{.002} & \secondmetric{0.076}{.002} & \bestmetric{2.649}{.009} & \metric{9.696}{.096} & \metric{1.751}{.062} \\
            
            \bottomrule
        \end{tabular}
    }
    
        \vspace{0.5em}
    \resizebox{0.8\linewidth}{!}
    {
        \begin{tabular}{c>{\ca}c>{\ca}c>{\ca}c>{\ca}c>{\ca}c>{\ca}c>{\ca}c}
            \toprule
            \multirow{2}{*}{Methods} & \multicolumn{3}{c}{R-Precision ↑} & \multirow{2}{*}{FID ↓} & \multirow{2}{*}{MM-Dist ↓} & \multirow{2}{*}{Diversity →} & \multirow{2}{*}{MultiModality ↑} \\
            & Top-1 & Top-2 & Top-3 &  &  &  &  \\
            \midrule
            Real motion & \metric{0.424}{.005} & \metric{0.649}{.006} & \metric{0.779}{.006} & \metric{0.031}{.004} & \metric{2.788}{.012} & \metric{11.08}{.097} & - \\ \midrule
            T2M~\cite{guo2022t2m} & \metric{0.370}{.005} & \metric{0.569}{.007} & \metric{0.693}{.007} & \metric{2.770}{.109} & \metric{3.401}{.008} & \metric{10.91}{.119} & \metric{1.482}{.065} \\
            AttT2M~\cite{zhong2023attt2m} & \metric{0.413}{.006} & \metric{0.632}{.006} & \metric{0.751}{.006} & \metric{0.870}{.039} & \metric{3.039}{.021} & \metric{10.96}{.123} & \bestmetric{2.281}{.047} \\
            ParCo~\cite{zou2024parco} & \metric{0.430}{.004} & \metric{0.649}{.007} & \metric{0.772}{.006} & \metric{0.453}{.027} & \metric{2.820}{.028} & \metric{10.95}{.094} & \metric{1.245}{.022} \\
            MoMask~\cite{guo2024momask} & \metric{0.433}{.007} & \metric{0.656}{.005} & \metric{0.781}{.005} & \secondmetric{0.204}{.011} & \secondmetric{2.779}{.022} & - & \metric{1.131}{.043} \\ \hdashline
            MDM~\cite{tevet2023mdm} & \metric{0.164}{.004} & \metric{0.291}{.004} & \metric{0.396}{.004} & \metric{0.497}{.021} & \metric{9.191}{.022} & \metric{10.847}{.109} & \metric{1.907}{.214} \\
            MLD~\cite{chen2023mld} & \metric{0.390}{.008} & \metric{0.609}{.008} & \metric{0.734}{.007} & \metric{0.404}{.027} & \metric{3.204}{.027} & \metric{10.80}{.117} & \secondmetric{2.192}{.071} \\
            MotionDiffuse~\cite{zhang2024motiondiffuse} & \metric{0.417}{.004} & \metric{0.621}{.004} & \metric{0.739}{.004} & \metric{1.954}{.062} & \metric{2.958}{.005} & \secondmetric{11.10}{.143} & \metric{0.730}{.013} \\
            ReMoDiffuse~\cite{zhang2023remodiffuse} & \metric{0.427}{.014} & \metric{0.641}{.004} & \metric{0.765}{.055} & \bestmetric{0.155}{.006} & \metric{2.814}{.012} & \metric{10.80}{.105} & \metric{1.239}{.028} \\ \midrule
            
            SALAD~(Ours) & \bestmetric{0.477}{.006} & \bestmetric{0.711}{.005} & \bestmetric{0.828}{.005} & \metric{0.296}{.012} & \bestmetric{2.585}{.016} & \bestmetric{11.097}{.095} & \metric{1.004}{.040} \\
            \bottomrule
        \end{tabular}
    }
    \vspace{-0.5em}
    \caption{
    Quantitative evaluation results on the test sets of HumanML3D~(top) and KIT-ML~(bottom).
    $\uparrow$ and $\downarrow$ denote that higher and lower values are better, respectively, while $\rightarrow$ denotes that the values closer to the real motion are better.
    Methods above the \dashuline{dotted line} are auto-regressive models based on VAE or VQ-VAE, while those below are diffusion-based generative models.
    \textcolor{red}{Red} and \textcolor{blue}{blue} colors indicate the best and the second best results, respectively.
    }
    \label{tab:quan}
    \vspace{-1em}
\end{table*}

\section{Experiments}
\label{sec:exp}
We conducted evaluations of our method on two widely used motion-language benchmarks: HumanML3D~\cite{guo2022t2m} and KIT-ML~\cite{plappert2016kit}.
HumanML3D dataset consists of 14,616 motion sequences with a variety of human actions, such as locomotions, sports, and acrobatics, along with 44,970 text descriptions in total.
KIT-ML dataset consists of 3,911 motion sequences along with 6,278 text descriptions.
Both datasets are augmented by mirroring, and divided into training, testing, and validation sets with the ratio of 0.8, 0.15, and 0.05.
Evaluation results are reported in the following sections.
Please refer to the supplementary material for additional quantitative experiment results with details, and the supplementary video for the animated results.

\begin{figure*}[t]
  \centering
   \includegraphics[width=0.95\linewidth]{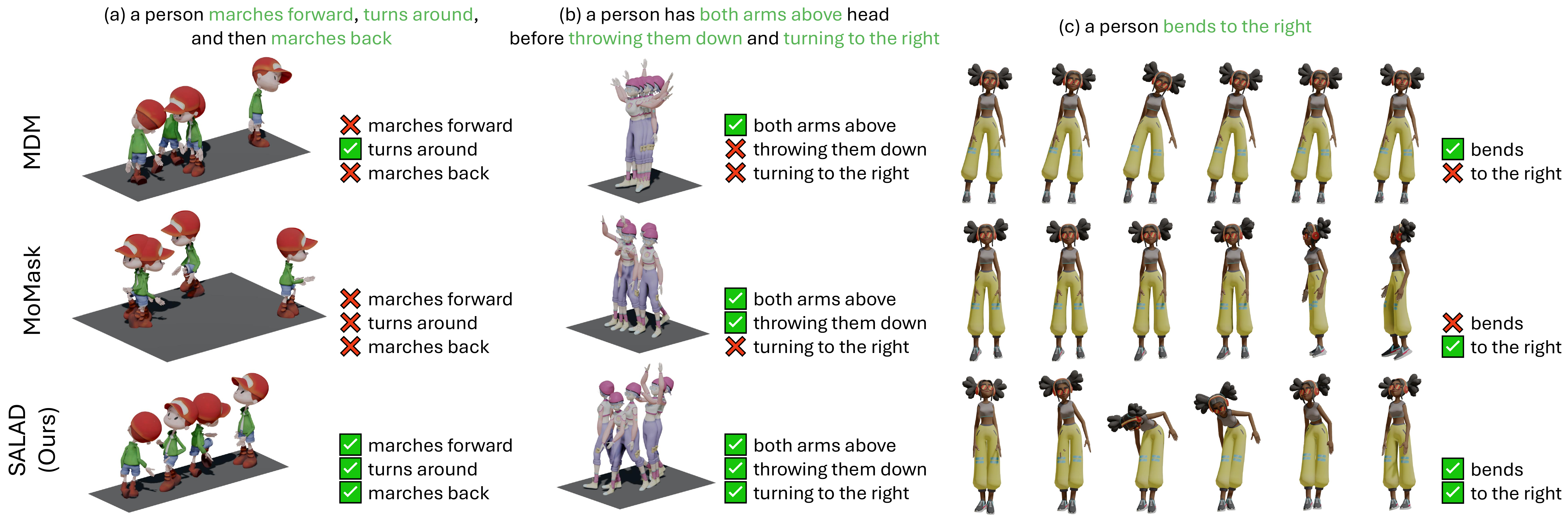}
   \vspace{-0.5em}
   \caption{Qualitative comparison of results generated by various methods, including MDM~\cite{tevet2023mdm}, MoMask~\cite{guo2024momask}, and our approach.}
   \label{fig:qual}
   \vspace{-1.5em}
\end{figure*}
\subsection{Implementation Details}
Our method was executed on a single NVIDIA V100 GPU.
We used the AdamW optimizer~\cite{loshchilov2018decoupled}, and the VAE and denoiser were trained for 50 and 500 epochs, respectively.
The VAE and denoiser were trained with a batch size of 64 for HumanML3D, which required 35 and 17 hours, respectively.
For the KIT-ML dataset, we used a batch size of 16, with training times of 20 hours for the VAE and 7 hours for the denoiser.
We trained the denoiser with 1000 diffusion steps, employing 50 steps for DDIM sampling during inference.
For the CFG weight, we set $w=7.5$ unless mentioned otherwise.

\subsection{Quantitative Evaluation}
\label{sec:quan}
For quantitative evaluations, we adopted metrics from T2M~\cite{guo2022t2m}.
First, \textit{FID} is used to evaluate the quality of generated motions by measuring the difference between the feature distributions of generated motions and those of real motions.
Additionally, \textit{R-precision} and \textit{MM-Dist} are used to evaluate the semantic alignment between the input text and the generated motions. 
\textit{Diversity}, which indicates the variance in generated motions, and \textit{MultiModality}, which assesses the diversity of motions generated from the same text description, are also adopted as secondary metrics. They are considered less important than the generation quality and alignment with the input text.

Following the practice of T2M~\cite{guo2022t2m}, each experiment was repeated 20 times on the test sets of HumanML3D and KIT-ML datasets, and the mean with a 95\% confidence interval is reported in~\cref{tab:quan}.
Notably, SALAD consistently achieved superior text-motion alignment across both datasets, regardless of the diffusion parametrization applied during denoiser training.
Moreover, SALAD maintained high generation quality, as indicated by the best or second-best FID scores among diffusion-based methods.
These results confirm that explicitly modeling interactions between skeletal joints, temporal frames, and textual words is effective in improving text-to-motion generation quality and fidelity to the input text.

\subsection{Qualitative Evaluation}
\label{sec:qual}
We compared the qualitative results of SALAD with those generated by MDM~\cite{tevet2023mdm} and MoMask~\cite{guo2024momask}, and the comparison results are shown in~\cref{fig:qual}.
Both MDM and MoMask reflected parts of the input text prompts, but they failed to handle all the descriptions in the text.
For example, as shown in the top of~\cref{fig:qual}-(a), MDM generated a character walking forward, turning, and walking back, but the locomotion style was not marching, despite the input prompt specifying \textit{"marches"}.
Similarly, in~\cref{fig:qual}-(b), MoMask accurately reflected the arm movements of raising above the head and throwing them down, but it missed the instruction to turn to the right.
In case of~\cref{fig:qual}-(c), both methods failed to perfectly reflect the input text despite its short length.
In contrast, SALAD consistently incorporated all the textual descriptions into the generated motions, regardless of the length and complexity of the text prompts.
This demonstrates its effectiveness in generating motions that are highly aligned with the input text.

\subsection{Ablation Study}
\label{sec:ablation}

\noindent\textbf{Effectiveness of Skeleton-aware Latent Space.}
To prove the effectiveness of using skeleton-aware latent space, we compared the performance of different VAE structures employed in MoMask~\cite{guo2024momask} and ParCo~\cite{zou2024parco} with ours.
Specifically, MoMask employs a VQ-VAE that compresses each pose into a single vector, while ParCo employs a part-aware VQ-VAE that processes different body parts independently.
Because the role of VAE is to compress the motion features into latent variables, we compared FID and mean per joint position error~(MPJPE) that assess the quality and accuracy of the reconstructed motion features, respectively.

As shown in~\cref{tab:abl-vae}, our skeleton-aware VAE outperformed baselines with a significant margin for both metrics.
Notably, our method required much fewer number of trainable parameters than the others.
This indicates that skeleton-aware latent space that decouples both dimensions and facilitates information exchange between adjacent components is effective in constructing meaningful motion latent space, which is a crucial prior to train the generator.

Furthermore, we conducted an ablation study that validates the effectiveness of the skeleto-temporal latent~(ST-Latent) and cross-attention, which are two core components of SALAD, on the text-to-motion generation performance.
For this experiment, we evaluated three variations: (i) replacing ST-Latent by training a conventional VAE that treats each pose as a single vector, (ii) replacing cross-attention with self-attention by concatenating the sentence-level CLIP feature to the motion features, and (iii) combining (i) and (ii).
As shown in~\cref{tab:stl_ca}, alteration of ST-Latent and cross-attention substantially degraded generation quality and text-motion alignment, as reflected in the higher FID and lower R-precision, respectively.
This demonstrates the importance of rich information exchange between text and motion by cross-attention within the skeleto-temporally structured latent space.
This indicates that the skeleto-temporally disentangled motion latent facilitates accurate text-motion interaction.

\begin{table}[t]
    \centering
    \resizebox{\linewidth}{!}
    {
        \begin{tabular}{c|>{\ca}c>{\ca}c>{\ca}c}
            \toprule
            Methods & Num of Params & FID ↓                & MPJPE ↓              \\ \midrule
            MoMask~\cite{guo2024momask}  & 19.44M        & \metric{0.020}{.000} & \metric{0.030}{.000} \\
            ParCo~\cite{zou2024parco}   & 6.35M         & \metric{0.021}{.000} & \metric{0.108}{.000} \\
            SALAD~(Ours)    & 0.16M         & \metric{0.003}{.000} & \metric{0.016}{.000} \\
            \bottomrule
        \end{tabular}
    }

    \vspace{-0.5em}
    \caption{Quantitative results on the quality and accuracy of reconstructed motion features of VAE models from different methods, along with the number of trainable parameters, measured on the test set of HumanML3D.}
    \label{tab:abl-vae}
    \vspace{-0.5em}
\end{table}
\begin{table}[t]
    \centering
    \resizebox{\linewidth}{!}
    {
        \begin{tabular}{c|>{\ca}c>{\ca}c}
            \toprule
            Method & R-Precision~(Top-3) ↑ & FID ↓ \\ \midrule
            Ours (Full model) & \metric{0.857}{.002} & \metric{0.076}{.002} \\
            w/o ST-Latent     & \metric{0.816}{.002} & \metric{0.433}{.006} \\
            w/o CrossAttn     & \metric{0.778}{.002} & \metric{0.274}{.007} \\
            w/o both          & \metric{0.752}{.002} & \metric{0.345}{.007} \\
            \bottomrule
        \end{tabular}
    }
    \vspace{-0.8em}
    \caption{Ablation studies on the VAE and denoiser.}
    \label{tab:stl_ca}
    \vspace{-1.7em}
\end{table}

\begin{figure}[t]
  \centering
    \vspace{-1em}
   \includegraphics[width=0.8\linewidth]{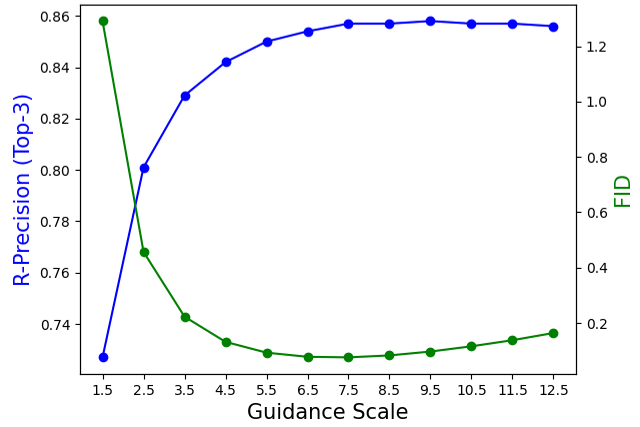}
   \vspace{-1em}
   \caption{Ablation study results on different CFG weights. Lower FID scores indicate better motion quality, and higher R-precision~(Top-3) scores indicate better semantic alignment.}
   \vspace{-1em}
   \label{fig:cfg_weights}
\end{figure}

\noindent\textbf{CFG Weights.}
To analyze the influence of CFG weights on generation results, we plotted FID and R-precision metric values in~\cref{fig:cfg_weights}.
The figure shows that low CFG weights led to suboptimal generation quality and text-motion alignment.
The performance of SALAD improved across both metrics as the weight values increased, but excessively high weight values resulted in a decreased performance for both metrics again.
Based on these findings, we determined that a value of $w=7.5$ produces the best balance between generation quality and alignment with the text prompts.

\subsection{Zero-shot Text-driven Motion Editing}
\label{sec:zero}
\begin{figure}[t]
  \centering
   \includegraphics[width=\linewidth]{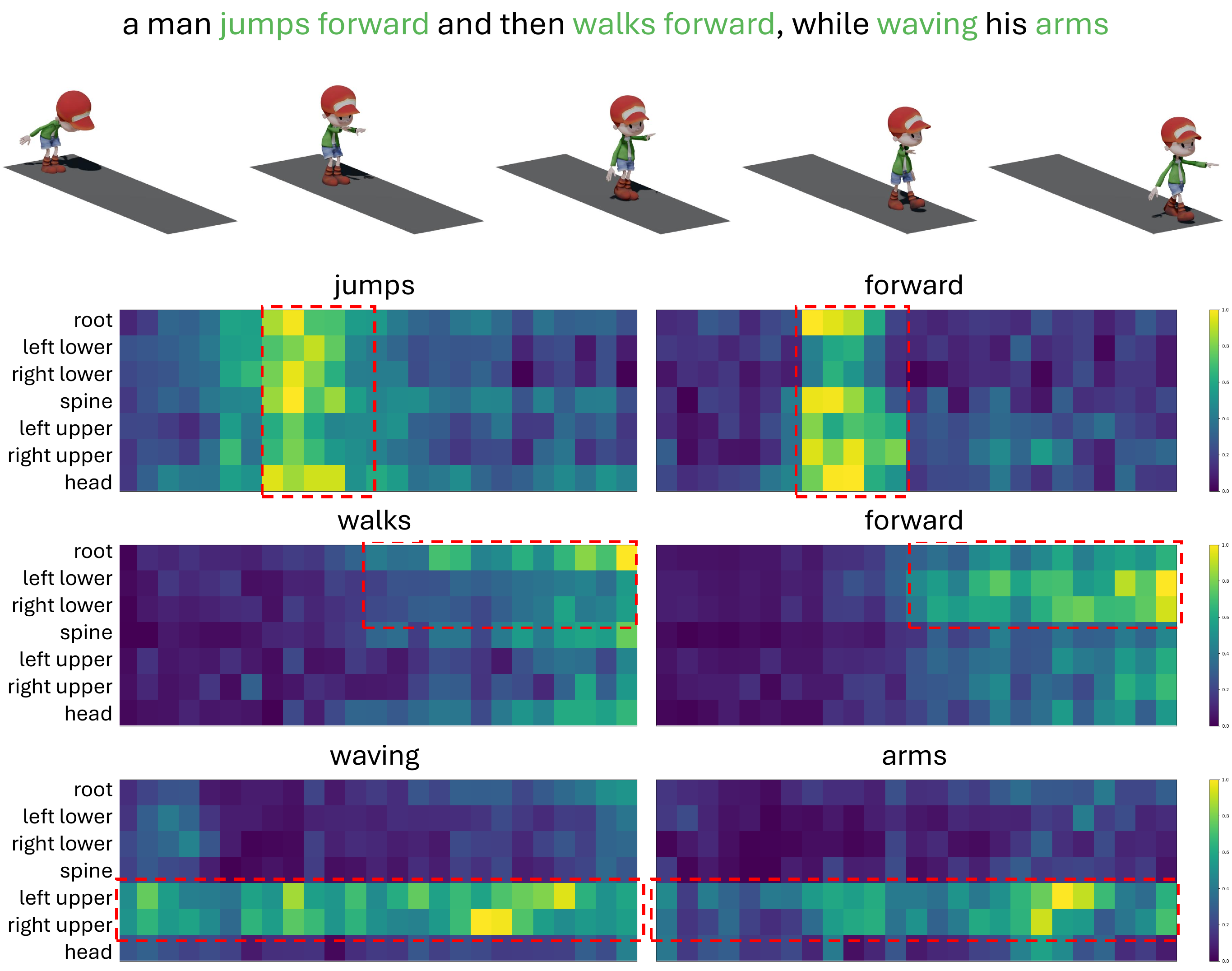}
   \vspace{-1.5em}
   \caption{Visualizations of cross-attention maps of SALAD. Each row corresponds to a specific body part, and each column represents temporal frames. Attention maps were computed by averaging the output attention maps across all transformer layers and all diffusion timesteps.}
   \label{fig:attn_map}
   \vspace{-1.5em}
\end{figure}

To demonstrate the capability of SALAD in capturing relationships between skeleto-temporal latents and text prompts, we visualized the cross-attention maps for each word in~\cref{fig:attn_map}.
The figure reveals that high attention values were assigned to proper frames and body parts that semantically correspond to each word.
For example, \textit{"jumps forward"} was activated for the entire body at early frames, \textit{"walks forward"} focused on the root and legs at later frames, and \textit{"waving arms"} was activated for the arms throughout the entire frames.
These results indicate that cross-attention maps produced during motion generation effectively capture the rich semantic relationships between text and motion.

\begin{figure*}[t]
  \centering
   \includegraphics[width=\linewidth]{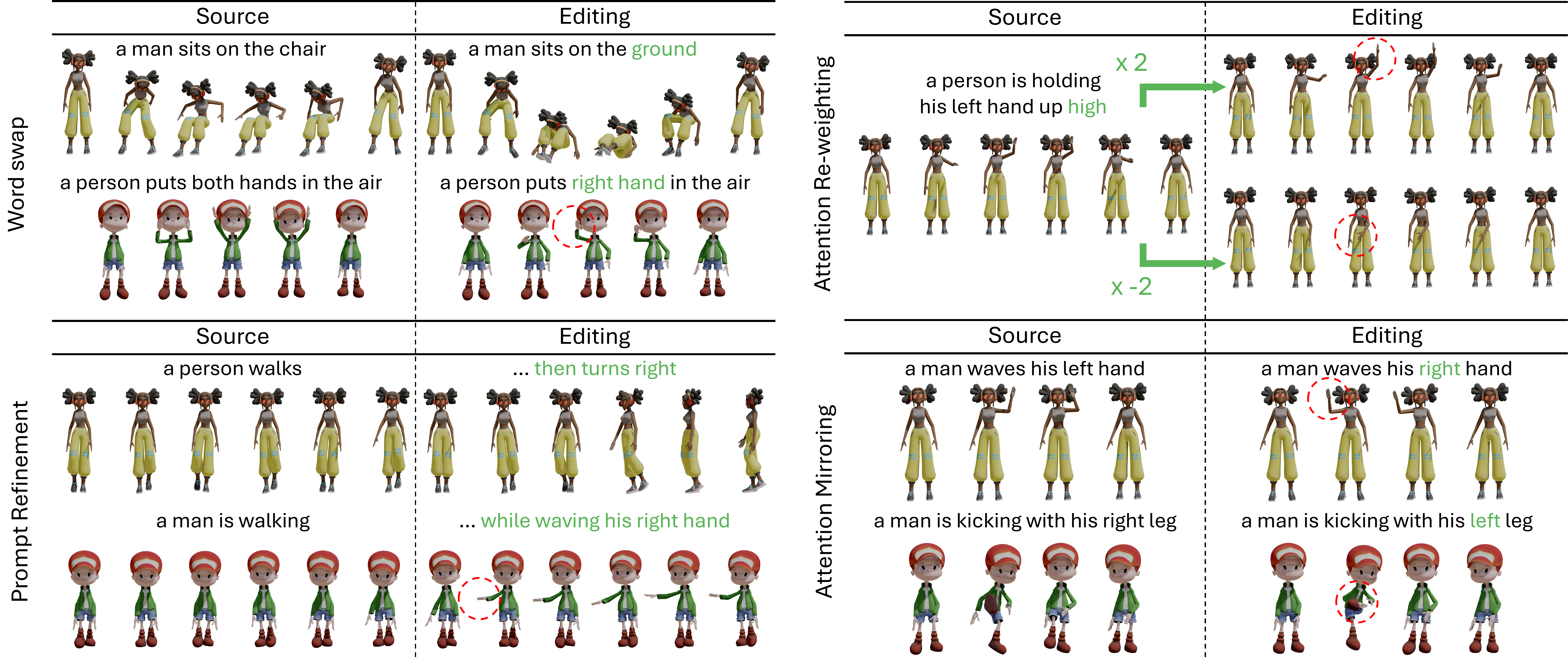}
   \vspace{-2em}
   \caption{Text-driven motion editing results via attention modulation using a pre-trained SALAD model. Texts  colored in green represent the editing instructions.}
   \label{fig:edit}
   \vspace{-1.5em}
\end{figure*}

Based on these observations, we demonstrate zero-shot text-driven motion editing results using a pre-trained SALAD model by modulating cross-attention maps, and the results are shown in~\cref{fig:edit}.
The edited motions align well with the given instructions while preserving the original movements, such as walking phases.
Notably, the changes for editing were localized to specific frames and body parts relevant to the editing prompt, while unaffected regions remain intact, without using explicit masking.
Furthermore, our SALAD model understands not only the high-level correspondence between text and motion but also its intensity and direction, which is demonstrated by different hand heights when modulating the word \textit{"high"} in the re-weighting case.
These results suggest that the explicit modeling of skeletal-temporal-textual relationships in SALAD enhances its understanding of text-driven motion generation, leading to effective learning representations and fine-grained control in both generation and editing.

\begin{table}[t]
    \centering
    \resizebox{\linewidth}{!}
    {
        \begin{tabular}{c|>{\ca}c>{\ca}c>{\ca}c}
            \toprule
            Methods & Preservation & Semantic Alignment & Overall Quality \\
            \midrule
            MDM~\cite{tevet2023mdm} & \metric{3.729}{.153}  & \metric{2.758}{.192} & \metric{3.196}{.147} \\
            MotionFix~\cite{athanasiou2024motionfix} & \metric{3.358}{.173}  & \metric{3.388}{.189} & \metric{3.421}{.166} \\
            SALAD~(Ours)    & \bestmetric{4.596}{.084}  & \bestmetric{4.654}{.087} & \bestmetric{4.596}{.083} \\
            \bottomrule
        \end{tabular}
    }
    \vspace{-0.5em}
    \caption{User study results. The \textcolor{Red}{red} color indicates the best result.}
    \label{tab:userstudy}
    \vspace{-1.5em}
\end{table}
To evalaute the perceptual quality of text-driven motion editing, we conducted a user study with 16 participants.
We compared SALAD against MDM~\cite{tevet2023mdm}, which edits motion by masking and regenerating specified body parts, and MotionFix~\cite{athanasiou2024motionfix}, a method specifically designed for motion editing.
Participants were shown with 15 videos per method and asked to rate them based on three criteria: (i) preservation of the original motion, (ii) semantic alignment with the target text, and (iii) overall quality.
Ratings were given on a 5-point Likert scale, with 5 indicating the best.

As shown in~\cref{tab:userstudy}, SALAD outperformed both baselines across all three aspects by a large margin.
Notably, despite not using explicit masking, SALAD achieved higher preservation scores.
Furthermore, while SALAD is not specifically designed for editing, it achieved higher preferences for semantic alignment and overall quality, highlighting the effectiveness of attention-based editing.
These results demonstrate that leveraging cross-attention maps enables high-quality and perceptually appealing text-driven motion editing without requiring additional training.
\vspace{-0.5em}
\section{Discussion}
While SALAD produced outperforming results in text-driven motion generation compared to the existing generative models, along with the capability of zero-shot text-driven motion editing, it has several limitations.
First, the numerical results for Diversity and MultiModality are limited, despite the strong capability of diffusion models to generate diverse outputs.
This can be due to the trade-off between high-quality with faithfulness to the text and diversity.
While we put the diversity as a second priority, finding a way to handle both could be an interesting approach.
Additionally, SALAD is restricted to generating a single person action with limited text and motion length.
Therefore, enabling multi-character interactions, crowd animations, generation from longer texts, and creating extended motion sequences could be an interesting direction for future work.
Furthermore, although our method does not directly address editing real motions, integrating diffusion inversion methods could enable text-driven editing of real motion sequences.
\section{Conclusion}
In this paper, we presented SALAD, a novel diffusion model framework for text-driven motion generation that models the complex interactions between skeletal joints, temporal frames, and textual words.
We first train a skeleto-temporal VAE that effectively summarizes motion features by decoupling skeletal and temporal dimensions, and we subsequently train a skeleto-temporal denoiser that enables efficient and expressive generation of motions conditioned on text input within this compact skeleto-temporal latent space.
Furthermore, we demonstrated the first text-driven zero-shot motion editing using a pre-trained SALAD model by modulating cross-attention maps during generation.
By leveraging the outperforming generative capacity compared to previous methods, along with the interpretable learning representations during generation, we anticipate that our work can benefit future research and applications in the context of text-driven motion generation and editing.

\section*{Acknowledgements}
We appreciate the anonymous reviewers for their invaluable discussion.
This work was supported by the National Research Foundation of Korea (NRF) grant funded by the Korea government (MSIT) (RS-2024-00333478).


{
    \small
    \bibliographystyle{ieeenat_fullname}
    \bibliography{main}
}
\clearpage
\setcounter{section}{0}
\setcounter{figure}{0}
\setcounter{table}{0}
\renewcommand{\thesection}{\Alph{section}} 


\twocolumn[{%
\renewcommand\twocolumn[1][]{#1}%
\maketitlesupplementary
\includegraphics[width=\linewidth]{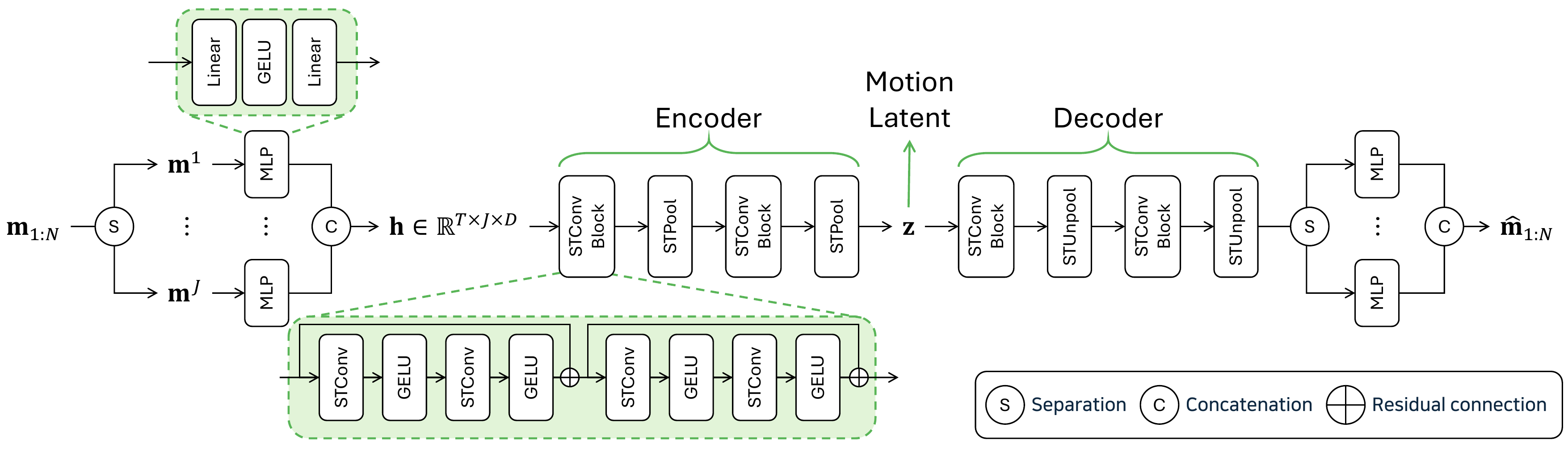}
\captionof{figure}{
Detailed architecture of the VAE network. Separation and concatenation of tensors are performed with respect to the joint dimension.
\vspace{1em}
}
\label{fig:supp-vae}
}]
\section{Network Architectures}
In this section, we provide a detailed explanation of the network architecture.
The code is available at {\href{https://github.com/seokhyeonhong/salad/}{github}.}

\subsection{Skeleto-temporal VAE}
We visualize the detailed architecture of the VAE network in Figure~\ref{fig:supp-vae}.
Given a motion sequence $\mathbf{m}$, we first decompose it into joint-wise features:
\begin{equation*}
    \mathbf{m} = \{ \mathbf{m}^{1}, \dots, \mathbf{m}^{J} \},
\end{equation*}
where $\mathbf{m}^{j} \in \mathbb{R}^{N \times D_{j}}$. Here, $N$ denotes the number of frames and $D_j$ represents the number of features of joint $j$, which varies depending on the joint.
Specifically, $D_j = 7$ for the root joint, as it includes 1-dimensional height, 2-dimensional translational velocity on the horizontal plane, 1-dimensional angular velocity around the up-axis, and 3-dimensional velocity.
For the foot and toe joints, $D_j = 13$, comprising 3-dimensional local positions and velocities, 6-dimensional local rotations, and a contact label.
For all the other joints, $D_j = 12$, excluding the contact label.
To match the dimension of each joint within the skeleton-aware latent space, we apply a joint-wise multi-layer perceptron~(MLP) to each $\mathbf{m}^{j}$, consisting of 2 linear layers and the Gaussian error linear unit~(GELU) activation function~\cite{hendrycks2016gelu} in the middle, yielding the joint-wise hidden latent variables $\mathbf{h}^{j}$.

\begin{figure}[t]
  \centering
   \includegraphics[width=\linewidth]{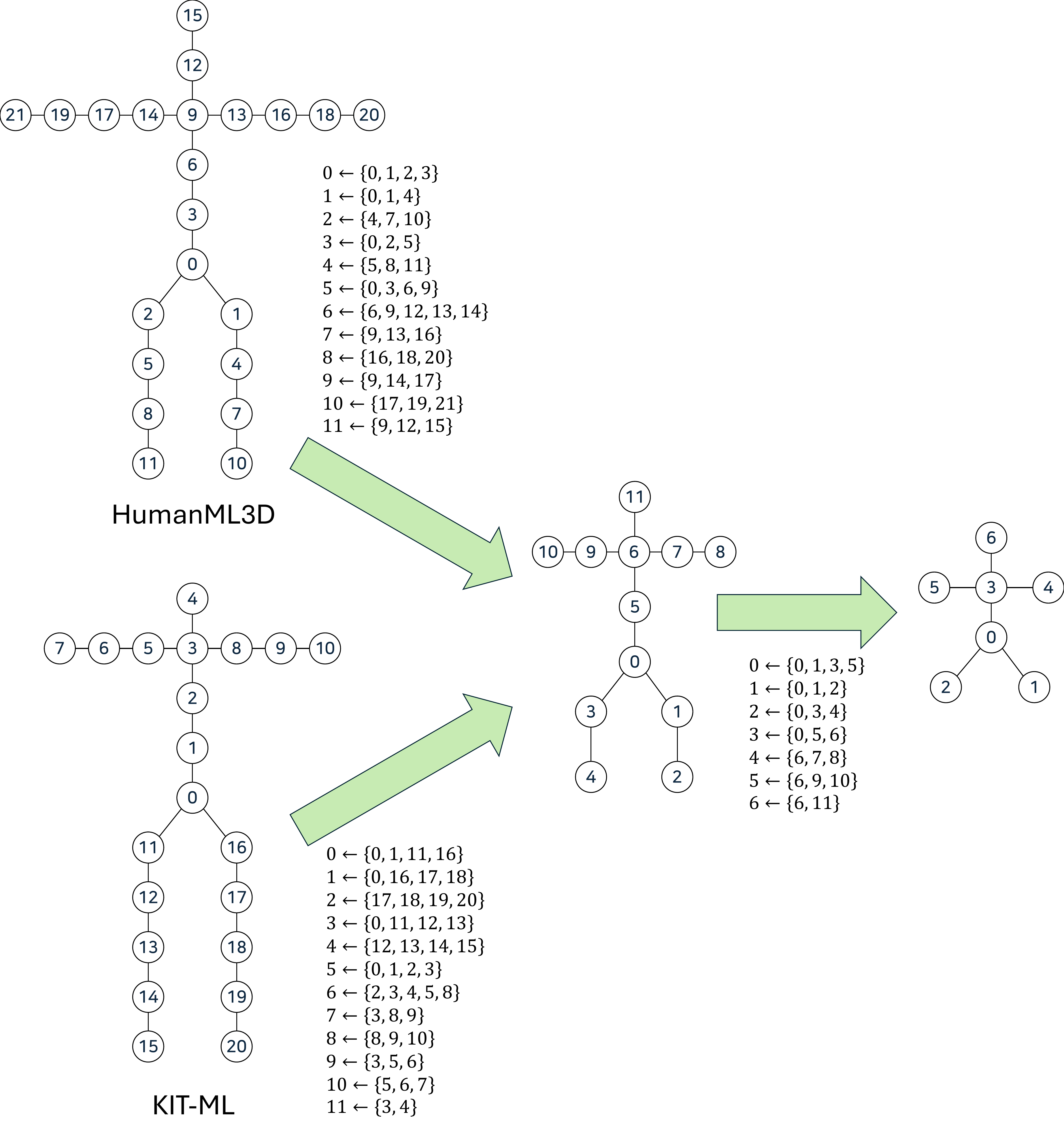}
   \caption{
   Illustration of the skeletal pooling process for the HumanML3D and KIT-ML datasets.
   The original skeleton~(left) is progressively abstracted by pooling adjacent joints~(middle and right).
   The notation $i \leftarrow \{ \}$ indicates the abstracted joint index and the set of original joints that are grouped together.
   The unpooling layers operate in the reverse order to restore the skeletal resolution.}
   \label{fig:skelpool}
   \vspace{-1em}
\end{figure}
As mentioned in the main paper, the skeleto-temporal convolution~(STConv) layers use a combination of a skeletal convolution~(SkelConv) and a temporal convolution~(TempConv).
The SkelConv layer is a graph convolution over the joint dimension, which is defined as follows:
\begin{equation*}
    \mathrm{SkelConv}(\mathbf{h}^{j}) \coloneq \boldsymbol{\Theta}_1(\mathbf{h}^{j}) + \frac{1}{\lvert \mathcal{N}(j) \rvert}\sum_{n \in \mathcal{N}(j)}\boldsymbol{\Theta}_2(\mathbf{h}^{n}),
\end{equation*}
where $\boldsymbol{\Theta}_{\{1, 2\}}$ represents a linear feed-forward layer, and $\mathcal{N}(j)$ denotes the indices of joints neighboring to joint $j$.
The TempConv is a 1D convolution layer with a kernel of size 3 and a stride of 1, and it is shared across all joints.
Additionally, we use residual connections within a stack of STConv layers.
For the activation function, we use the GELU function at the end of each STConv layer.

For the skeleto-temporal pooling~(STPool) layers, we apply the average pooling across both skeletal and temporal dimensions.
Specifically, temporal features are pooled with a kernel of size 2 and a stride of 2, while skeletal pooling reduces the number of joints by summarizing adjacent joints.
In contrast, the skeleto-temporal unpooling~(STUnpool) layers perform the inverse operation of STPool.
To increase the skeletal resolution, we recreate unpooled joints by summing the features from the corresponding pooled joints.
The temporal resolution is increased by upsampling the features along the temporal dimension using linear interpolation.
The downsampling and upsampling for each dataset is visualized in Figure~\ref{fig:skelpool}.

\subsection{Skeleto-temporal Denoiser}
In this section, we provide the implementation details for each component of the skeleto-temporal denoiser.
\noindent{\textbf{Positional Embedding.}}
We incorporate order information of $\zlt$ for both temporal and skeletal dimensions by applying positional embedding.
Specifically, we first compute the positional embedding $\mathbf{e} \in \mathbb{R}^{TJ \times D}$ using the sinusoidal positional embedding method~\cite{vaswani2017attention}.
We then reshape this tensor to $\mathbb{R}^{T \times J \times D}$, and add it to the motion latent $\zlt$.

\noindent{\textbf{FiLM.}}
To feed the diffusion timestep information to the denoiser, we employ the FiLM operator combined with an MLP layer, which has shown impressive performance in motion diffusion models~\cite{tseng2023edge, zhang2024motiondiffuse}.
Specifically, the diffusion timestep $t$ is passed through a sinusoidal positional embedding, followed by an MLP layer, resulting in a scale factor $\boldsymbol{\gamma}_t$ and shift factor $\boldsymbol{\beta}_t$, both $D$-dimensional vectors used to modulate the output of network modules:
\begin{equation}
    \mathrm{FiLM}(\mathbf{z}, t) = \boldsymbol{\gamma}_t \odot \mathbf{z} + \boldsymbol{\beta}_t,
\end{equation}
where $\gamma_t$ and $\beta_t$ are produced for each of attention block and feed-forward network in each layer.
For brevity, we omit $t$ for the $\mathrm{FiLM}$ in the following sections.

\noindent{\textbf{Feed-forward Network.}}
To enhance the non-linearity capacity of the model, we employ an FFN module combined with FiLM:
\begin{gather}
    \vfunc{FFN}{\zlt} \coloneq \vfunc{MLP}{\vfunc{GELU}{\vfunc{MLP}{\vfunc{LN}{\zlt}}}}, \\
    \mathbf{z}^{l+1}_t \leftarrow \zlt + \vfunc{FiLM}{\vfunc{FFN}{\zlt}},
\end{gather}
where GELU represents the Gaussian error linear unit activation function~\cite{hendrycks2016gelu}.
Notably, FFN module produces the output of the $l$-th transformer layer, which we denote as $\mathbf{z}^{l+1}_t$, and it is used as an input of the $(l+1)$-th layer.

\subsection{Hyperparameters}
For training the VAE, we set $\lambda_\mathrm{pos}=0.5$, $\lambda_\mathrm{vel}=0.5$, and $\lambda_\mathrm{kl}=0.02$.
We also used a learning rate scheduler for training both the VAE and denoiser, which linearly increased the learning rate from 0 to 0.0002 over the first 2000 steps, then decayed it by a factor of 0.1 at 150,000 and 250,000 iterations for the VAE, and at 50,000 iterations for the denoiser.
The latent dimensions for modules of the skeleto-temporal VAE was set to 32, while the denoiser used a latent dimension of 256.

We used a scaled linear scheduler for the noise schedule during denoiser training, similar to the approach used in Stable Diffusion~\cite{rombach2022high}.
Specifically, $\beta_t$ at diffusion timestep $t$ is defined as follows:
\begin{equation}
    \beta_t = \left( \sqrt{\beta_1} + \frac{t-1}{T-1}\cdot\left(\sqrt{\beta_T} - \sqrt{\beta_1} \right) \right)^2,
\end{equation}
where we set $\beta_1 = 0.00085$, $\beta_T = 0.012$, and $T=1000$ following the default setting of Stable Diffusion.
This scheduler allows for a smooth transition between noise levels, ensuring effective denoising throughout the entire process.
\newcommand{\dm}{\mathrm{Denoise}}

\section{Zero-shot Editing via Attention Modulation}
To enable zero-shot editing by modulating cross-attention maps between motion and text, we adopt the editing procedure introduced by Prompt-to-Prompt~\cite{hertz2022prompt2prompt}, which is depicted in Algorithm~\ref{algo:editing}.
Let $\dm(\mathbf{z}_t, t, c)$ represent a single denoising step at diffusion timestep $t$ using a pre-trained SALAD model conditioned on text prompt $c$, producing the denoised latent $\mathbf{z}_{t-1}$ and cross-attention map $\mathbf{M}_t$.
Additionally, $\dm(\mathbf{z}_t, t, c)\{ \mathbf{M} \gets \hat{\mathbf{M}} \}$ denotes a denoising step where the original attention map $\mathbf{M}$ is replaced by a modified attention map $\hat{\mathbf{M}}$, while the value $\mathbf{V}$ is computed using the target prompt $c^*$.
The function $\mathrm{Edit}(\mathbf{M}_t, \mathbf{M}_t^*, t)$ is a general editing function that modulates two attention maps based on the diffusion timestep $t$, which will be elaborated in the following sections.

\begin{algorithm}[t]
    \caption{Zero-shot editing by cross-attention modulation}
    \begin{algorithmic}[1]
        \STATE \textbf{Input:} A source prompt $c$ and a target prompt $c^*$.
        \STATE \textbf{Output:} A source motion $\mathbf{z}_0$ and an edited motion $\mathbf{z}_0^*$.
        \STATE $\mathbf{z}_T \sim \mathcal{N}(0, I)$ a unit Gaussian random variable;
        \STATE $\mathbf{z}_T^* \gets \mathbf{z}_T$;
        \FOR{$t = T, T - 1, \ldots, 1$}
            \STATE $\mathbf{z}_{t-1}, \mathbf{M}_t \gets \dm(\mathbf{z}_t, t, c)$;
            \STATE $\mathbf{z}_{t-1}^*, \mathbf{M}_t^* \gets \dm(\mathbf{z}^*_t, t, c^*)$;
            \STATE $\hat{\mathbf{M}}_t \gets \mathrm{Edit}(\mathbf{M}_t, \mathbf{M}_t^*, t)$;
            \STATE $\mathbf{z}^*_{t-1} \gets \dm(\mathbf{z}_t^*, t, c^*) \{\mathbf{M} \gets \hat{\mathbf{M}}_t\}$;
        \ENDFOR
        \RETURN $(\mathbf{z}_0, \mathbf{z}^*_0)$
    \end{algorithmic}
    \label{algo:editing}
\end{algorithm}

\noindent\textbf{Word Swap.}
The objective of word swap is to incorporate descriptions from the target prompt while maintaining movements from the source prompt.
To this end, the source prompt is used during early denoising steps, and the target prompt is used in the later steps:
\begin{equation*}
    \mathrm{Edit}(\mathbf{M}_t, \mathbf{M}_t^*, t) \coloneq
    \begin{cases}
        \mathbf{M}_t^* & \text{if $t < \tau$} \\
        \mathbf{M}_t   & \text{otherwise},    \\
    \end{cases}
\end{equation*}
where $\tau$ is a hyperparameter that determines when to switch from the source prompt to the target prompt during the denoising process.
Because the overall composition is established in the early steps, we guide the overall structure by injecting the attention maps from the source prompt during the early steps, and we use the attention maps of the target prompt to refine details later.
Although the optimal value of $\tau$ can vary depending on the specific motion and text inputs, we found that $\tau = 0.8T$ generally yields good results.

\noindent\textbf{Prompt Refinement.}
In this case, the target prompt is created by appending new tokens to the source text to add additional details.
We first obtain the cross-attention map of the target text prompt $\mathbf{M}^*$, and then overwrite the attention values for existing tokens with those from the original cross-attention map $\mathbf{M}$ to preserve the common information between the two prompts:
\begin{equation*}
    (\mathrm{Edit}(\mathbf{M}_t, \mathbf{M}_t^*, t))_{i, j, k} \coloneq
    \begin{cases}
        (\mathbf{M}_t^*)_{i, j, k}            & \text{if $\mathcal{A}(j) = -1$} \\
        (\mathbf{M}_t)_{i, j, \mathcal{A}(k)} & \text{otherwise}, \\
    \end{cases}
\end{equation*}
where $i$, $j$, and $k$ denote the indices of the skeletal, temporal, and text tokens, respectively. The function $\mathcal{A}(\cdot)$ is an alignment function that maps a token index from the target prompt $c^*$ to its corresponding index in $c$ if one exists, or returns $-1$ otherwise.

\noindent\textbf{Attention Re-weighting.}
This editing case involves amplifying or reducing the attention weights associated with specific word tokens to influence the resulting motions.
Given a word token $k^*$ and a scaling parameter $s$, we adjust the attention values as follows:
\begin{equation*}
    (\mathrm{Edit}(\mathbf{M}_t, \mathbf{M}_t^*, t))_{i, j, k} \coloneq
    \begin{cases}
        s\cdot(\mathbf{M}_t)_{i, j, k} & \text{if $k=k^*$} \\
        (\mathbf{M}_t)_{i, j, k}       & \text{otherwise}. \\
    \end{cases}
\end{equation*}
We empirically found that setting $s$ within the range of $[-3, 3]$ yields reasonable editing results.
Notably, using negative weights can generate semantically opposite outcomes, indicating that the cross-attention maps in SALAD capture a sophisticated understanding of the high-level relationships between text and motion.

\noindent\textbf{Attention Mirroring.}
In this case, cross-attention values between counterpart body parts, such as the left and right arms, are swapped:
\begin{equation*}
    (\mathrm{Edit}(\mathbf{M}_t, \mathbf{M}_t^*, t))_{i, j, k} \coloneq (\mathbf{M}_t)_{i, \mathcal{C}(j), k}
\end{equation*}
where $\mathcal{C}(j)$ outputs the joint index of the counterpart if it exists, or $j$ otherwise.
This approach effectively mirrors the motion without needing to compute a new cross-attention map $\hat{\mathbf{M}}$ corresponding to an editing text prompt.

\section{Additional Experiment Results}
\subsection{Cross-attention Maps}
\begin{figure*}[t]
  \centering
   \includegraphics[width=\linewidth]{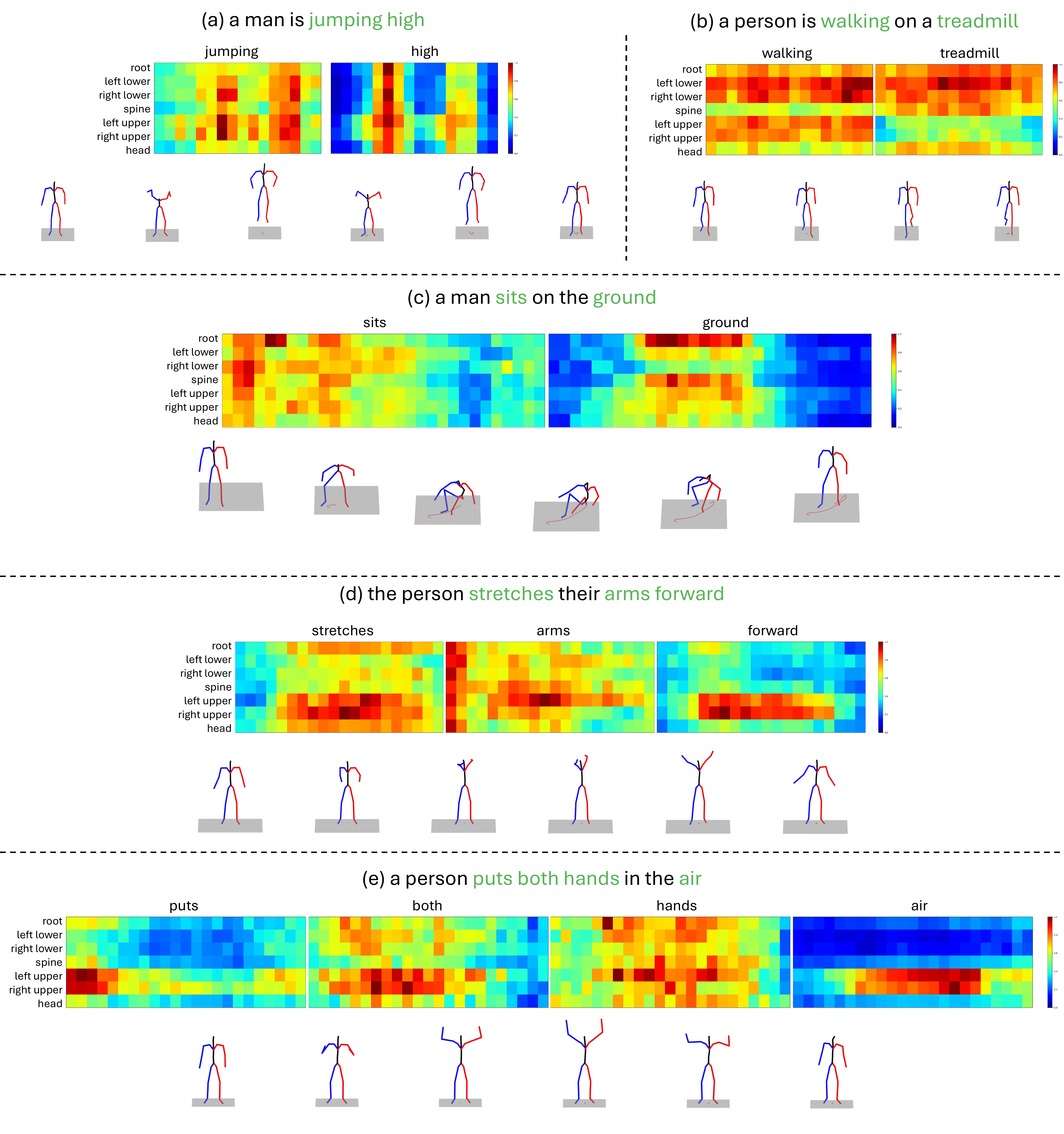}
   \caption{Additional visualizations of cross-attention maps between text and motion. Each row corresponds to a specific body part, and each column represents temporal frames.}
   \label{fig:supp-attn-map}
   \vspace{-1em}
\end{figure*}
We present additional visualizations of cross-attention maps between the input text and generated motions in Figure~\ref{fig:supp-attn-map}.
Overall, the attention maps consistently captured the relationships between the text and motion as reflected in the generated results.
In Figure~\ref{fig:supp-attn-map}-(a), attention peaks for \textit{jumping} and \textit{high} occurred twice along the temporal dimension, corresponding to the character jumping twice, indicating that the attention map effectively captures the relationship between frames and words.
Additionally, as shown in Figure~\ref{fig:supp-attn-map}-(b), the words \textit{walking} and \textit{treadmill} were associated with the lower body parts, showing consistently high weights along the frames, demonstrating that the attention map captures the relationship between body parts and words.
Figure~\ref{fig:supp-attn-map}-(c), (d), and (e) further show that the cross-attention maps capture the relationship between skeleto-temporal features and each word, conveying which body part should to be activated, at which timing, and where to act, such as \textit{sits ... ground}, \textit{stretches arms ... forward}, and \textit{puts both hands ... air}.
Overall, these results demonstrate that cross-attention maps effectively capture the relationship between skeleto-temporal features and each word in the textual descriptions.

\subsection{Classifier-free Guidance Weights}
\begin{table*}[t]
    \centering
    \begin{tabular}{c>{\ca}c>{\ca}c>{\ca}c>{\ca}c>{\ca}c>{\ca}c}
        \toprule
        \multirow{2}{*}{$w$} & \multicolumn{3}{c}{R-Precision ↑} & \multirow{2}{*}{FID ↓} & \multirow{2}{*}{MM-Dist ↓} & \multirow{2}{*}{Diversity →} \\
        & Top-1 & Top-2 & Top-3 &  &  & \\
        \midrule
        Real motion & \metric{0.511}{.003} & \metric{0.703}{.003} & \metric{0.797}{.002} & \metric{0.002}{.000} & \metric{2.974}{.008} & \metric{9.503}{.065} \\
        \midrule
        1.5 & \metric{0.438}{.003} & \metric{0.622}{.002} & \metric{0.727}{.002} & \metric{1.291}{.020} & \metric{3.411}{.011} & \metric{9.298}{.085} \\
        2.5 & \metric{0.517}{.003} & \metric{0.705}{.002} & \metric{0.801}{.002} & \metric{0.457}{.011} & \metric{2.945}{.011} & \metric{9.616}{.089} \\
        3.5 & \metric{0.548}{.003} & \metric{0.738}{.002} & \metric{0.829}{.002} & \metric{0.223}{.006} & \metric{2.782}{.009} & \metric{9.711}{.092} \\
        4.5 & \metric{0.563}{.003} & \metric{0.755}{.002} & \metric{0.842}{.002} & \metric{0.132}{.004} & \metric{2.711}{.008} & \metric{9.737}{.090} \\
        5.5 & \metric{0.573}{.003} & \metric{0.763}{.002} & \metric{0.850}{.002} & \metric{0.093}{.003} & \metric{2.674}{.008} & \metric{9.741}{.094} \\
        6.5 & \secondmetric{0.578}{.003} & \metric{0.767}{.002} & \metric{0.854}{.002} & \secondmetric{0.078}{.003} & \metric{2.656}{.009} & \metric{9.722}{.095} \\
        7.5~(default) & \bestmetric{0.581}{.003} & \metric{0.769}{.003} & \secondmetric{0.857}{.002} & \bestmetric{0.076}{.002} & \bestmetric{2.649}{.009} & \metric{9.696}{.096} \\
        8.5 & \bestmetric{0.581}{.003} & \secondmetric{0.770}{.002} & \secondmetric{0.857}{.002} & \metric{0.083}{.002} & \secondmetric{2.650}{.009} & \metric{9.669}{.094} \\
        9.5 & \bestmetric{0.581}{.003} & \bestmetric{0.771}{.002} & \bestmetric{0.858}{.002} & \metric{0.097}{.003} & \metric{2.655}{.009} & \metric{9.638}{.093} \\
        10.5 & \bestmetric{0.581}{.003} & \secondmetric{0.770}{.002} & \secondmetric{0.857}{.001} & \metric{0.116}{.003} & \metric{2.663}{.008} & \secondmetric{9.608}{.094} \\
        11.5 & \bestmetric{0.581}{.002} & \metric{0.769}{.002} & \metric{0.857}{.002} & \metric{0.138}{.003} & \metric{2.673}{.008} & \bestmetric{9.577}{.093} \\
        12.5 & \secondmetric{0.578}{.002} & \metric{0.768}{.002} & \metric{0.856}{.002} & \metric{0.164}{.003} & \metric{2.686}{.008} & \metric{9.657}{.091} \\
        \bottomrule
    \end{tabular}

    \vspace{0.5em}
    
    \begin{tabular}{c>{\ca}c>{\ca}c>{\ca}c>{\ca}c>{\ca}c>{\ca}c}
        \toprule
        \multirow{2}{*}{CFG Weights} & \multicolumn{3}{c}{R-Precision ↑} & \multirow{2}{*}{FID ↓} & \multirow{2}{*}{MM-Dist ↓} & \multirow{2}{*}{Diversity →} \\
        & Top-1 & Top-2 & Top-3 & & & \\
        \midrule
        Real motion & \metric{0.424}{.005} & \metric{0.649}{.006} & \metric{0.779}{.006} & \metric{0.031}{.004} & \metric{2.788}{.012} & \metric{11.08}{.097} \\
        \midrule
        1.5 & \metric{0.439}{.005} & \metric{0.656}{.006} & \metric{0.774}{.007} & \metric{0.837}{.033} & \metric{2.755}{.018} & \metric{11.033}{.115} \\
        2.5 & \metric{0.471}{.006} & \metric{0.691}{.005} & \metric{0.809}{.005} & \metric{0.489}{.017} & \metric{2.593}{.013} & \metric{11.085}{.118} \\
        3.5 & \metric{0.480}{.006} & \metric{0.706}{.005} & \metric{0.818}{.006} & \metric{0.396}{.013} & \secondmetric{2.562}{.013} & \metric{11.086}{.120} \\
        4.5 & \bestmetric{0.484}{.006} & \secondmetric{0.709}{.005} & \metric{0.823}{.006} & \metric{0.352}{.011} & \bestmetric{2.559}{.015} & \metric{11.083}{.122} \\
        5.5 & \bestmetric{0.484}{.006} & \bestmetric{0.710}{.005} & \secondmetric{0.824}{.006} & \metric{0.324}{.009} & \metric{2.571}{.016} & \secondmetric{11.078}{.122} \\
        6.5 & \secondmetric{0.481}{.007} & \metric{0.708}{.006} & \secondmetric{0.824}{.006} & \metric{0.308}{.009} & \metric{2.588}{.016} & \metric{11.077}{.119} \\
        7.5~(default) & \metric{0.480}{.007} & \metric{0.707}{.006} & \bestmetric{0.826}{.006} & \metric{0.298}{.009} & \metric{2.608}{.015} & \bestmetric{11.079}{.119} \\
        8.5 & \metric{0.476}{.008} & \metric{0.705}{.005} & \metric{0.823}{.006} & \bestmetric{0.294}{.008} & \metric{2.630}{.015} & \metric{11.083}{.120} \\
        9.5 & \metric{0.477}{.007} & \metric{0.704}{.005} & \metric{0.822}{.005} & \bestmetric{0.294}{.008} & \metric{2.651}{.015} & \metric{11.089}{.120} \\
        10.5 & \metric{0.474}{.006} & \metric{0.703}{.005} & \metric{0.820}{.006} & \secondmetric{0.297}{.009} & \metric{2.674}{.015} & \metric{11.096}{.121} \\
        11.5 & \metric{0.472}{.006} & \metric{0.702}{.006} & \metric{0.818}{.006} & \metric{0.303}{.009} & \metric{2.697}{.015} & \metric{11.102}{.121} \\
        12.5 & \metric{0.471}{.006} & \metric{0.697}{.006} & \metric{0.815}{.006} & \metric{0.311}{.010} & \metric{2.719}{.015} & \metric{11.108}{.122} \\
        \bottomrule
    \end{tabular}

    \vspace{-0.5em}
    \caption{
    Quantitative evaluation results with different CFG weight values on the test sets of HumanML3D~(top) and KIT-ML~(bottom).
    $\uparrow$ and $\downarrow$ denote that higher and lower values are better, respectively, while $\rightarrow$ denotes that the values closer to the real motion are better.
    \textcolor{red}{Red} and \textcolor{blue}{blue} colors indicate the best and the second best results, respectively.
    }
    \label{tab:cfg-scale}
    \vspace{-1em}
\end{table*}
We present the quantitative results for different classifier-free guidance~(CFG) weight values in Table~\ref{tab:cfg-scale}.
As mentioned in the main paper, the performance of SALAD improved across both metrics as the weight values increased, but excessively high weight values resulted in a decline in performance for both metrics.
Notably, R-precision exhibited marginal differences for CFG weight values greater than or equal to 7.5, remaining within statistically equivalent ranges.
In contrast, for FID and MM-Dist on the HumanML3D dataset, the default setting $w=7.5$ yielded the best results.
Similarly, in the KIT-ML dataset, $w=7.5$ provided a balance between the quality and text-motion alignment.
For Diversity, $w=1.5$ produced results significantly inferior to the ground truth in both datasets, but the performance progressively improved as the CFG weight values increased.
However, the optimal point at which the best Diversity is achieved was different between datasets: $w=11.5$ for HumanML3D and $w=7.5$ for KIT-ML.

\subsection{Diffusion Parametrization}
To demonstrate the effectiveness of $\mathbf{v}$-prediction parametrization, we compared the results of SALAD models trained with different parametrizations, as shown in the last 3 rows of Table~\ref{tab:quan}.
While any variation of SALAD produced comparable or outperforming results compared to previous methods, $\velocity$-prediction consistently produced more stable results than the other alternatives across different datasets.
Specifically, on the HumanML3D dataset, $\sample$- and $\velocity$-prediction yielded statistically equivalent results for R-precision within the confidence range, but $\velocity$-prediction achieved superior results on FID and MM-Dist compared to $\sample$-prediction, while $\noise$-prediction consistently produced inferior results.
On the KIT-ML dataset, $\noise$-prediction scored better on FID compared to $\sample$-prediction, while both achieved similar results in terms of the text-motion alignment.
Also on this dataset, $\velocity$-prediction significantly outperformed both alternatives.
These results demonstrate the effectiveness of $\velocity$-prediction in enhancing stability and robustness in text-to-motion generation using SALAD.
\begin{table}[t]
    \centering
    \begin{tabular}{c|>{\ca}c>{\ca}c}
        \toprule
        Parametrization & R-Precision~(Top-3) ↑ & FID ↓ \\ \midrule
        $\sample$-prediction   & \metric{0.860}{.002} & \metric{0.111}{.003} \\
        $\noise$-prediction    & \metric{0.803}{.002} & \metric{0.257}{.008} \\
        $\velocity$-prediction & \metric{0.857}{.002} & \metric{0.076}{.002} \\
        \bottomrule
    \end{tabular}
    \caption{Quantitative results on different diffusion parametrizations.}
    \vspace{-1em}
    \label{tab:abl-diff-param}
\end{table}

\subsection{FiLM Layers}
While FiLM has been adopted in several motion diffusion models~\cite{tseng2023edge, zhang2024motiondiffuse}, its effectiveness in motion generation has yet to be fully evaluated.
Therefore, we compared the performance of SALAD with and without FiLM, as shown in Table~\ref{tab:abl-film}.
For the model without FiLM layers, we injected diffusion timestep information to $\zlt$ by directly adding the positional embedding of diffusion timestep $t$ to it.
Across both datasets, FiLM layers improved both text-motion alignment and generation quality, indicating that explicit modulation of intermediate representations effectively produces high-quality and text-faithful results.
\begin{table}[t]
    \centering
    \begin{tabular}{c|>{\ca}c>{\ca}c}
        \toprule
        Method & R-Precision~(Top-3) ↑ & FID ↓ \\ \midrule
        with FiLM & \metric{0.857}{.002} & \metric{0.076}{.002} \\
        without FiLM & \metric{0.843}{.002} & \metric{0.087}{.003} \\
        \bottomrule
    \end{tabular}

    \vspace{0.5em}
    
    \begin{tabular}{c|>{\ca}c>{\ca}c}
        \toprule
        Method & R-Precision~(Top-3) ↑ & FID ↓ \\ \midrule
        with FiLM & \metric{0.828}{.005} & \metric{0.296}{.012} \\
        without FiLM & \metric{0.803}{.006} & \metric{0.311}{.013} \\
        \bottomrule
    \end{tabular}
    \vspace{-0.5em}
    \caption{Ablation results showing the effect of FiLM layers on the test sets of HumanML3D (top) and KIT-ML (bottom).}
    \label{tab:abl-film}
\end{table}

\subsection{Loss Terms of VAE}
To validate the effectiveness of the auxiliary loss terms in the VAE, including $\mathcal{L}_\mathrm{pos}$ and $\mathcal{L}_\mathrm{vel}$, we evaluated FID and MPJPE while ablating these loss terms, as shown in~\cref{tab:abl-vae-loss}.
Removing $\mathcal{L}_\mathrm{pos}$ had a negative effect in both metrics, while $\mathcal{L}_\mathrm{vel}$ had no effect on MPJPE but led to a more significant degradation in FID compared to ablating $\mathcal{L}_\mathrm{pos}$.
Furthermore, removing both loss terms resulted in the worst performance across all metrics.
These results demonstrate the importance of auxiliary losses on joint positions and velocities in refining the latent space, contributing to improving motion quality and reliability.
\begin{table}[t]
    \centering
    \begin{tabular}{c|>{\ca}c>{\ca}c}
        \toprule
        Method & FID ↓ & MPJPE ↓ \\ \midrule
        Full model & \metric{0.003}{.000} & \metric{0.016}{.000} \\
        without $\mathcal{L}_\mathrm{pos}$ & \metric{0.005}{.000} & \metric{0.023}{.000} \\
        without $\mathcal{L}_\mathrm{vel}$ & \metric{0.009}{.000} & \metric{0.016}{.000} \\
        without both & \metric{0.012}{.000} & \metric{0.024}{.000} \\
        \bottomrule
    \end{tabular}
    \caption{Ablation results showing the effect of FiLM layers on the test sets of HumanML3D (top) and KIT-ML (bottom).}
    \label{tab:abl-vae-loss}
\end{table} 

\end{document}